\RequirePackage{amsmath}
\documentclass[runningheads]{llncs}
\usepackage[T1]{fontenc}
\usepackage{xcolor}
\usepackage{graphicx}
\usepackage{subfigure}
\usepackage{algorithmic}
\usepackage{amssymb,amsfonts}
\usepackage{makecell}
\usepackage[misc]{ifsym}

\begin{document}

\title{A Visual Interpretation-Based Self-Improved Classification System Using Virtual Adversarial Training
\thanks{Supported by KAKENHI (20K11830) Japan.}
}
\titlerunning{A Visual Interpretation-Based Self-Improved Classification System}
%-------------------------------------------------------
% \iffalse

\author{Shuai Jiang\inst{1}$^{(\textrm{\Letter})}$ \and
Sayaka Kamei\inst{1} \and
Chen Li\inst{2} \and 
Shengzhe Hou\inst{3} \and \\
Yasuhiko Morimoto\inst{1}
}
\authorrunning{S. Jiang et al.}

\institute{
Graduate School of Advanced Science and Engineering, Hiroshima University, Japan.\\
%\email{\{jiangshuai, s10kamei, morimo\}@hiroshima-u.ac.jp}
\and
Graduate School of Informatics, Nagoya University, Japan.\\
%\email{li.chen.z2@a.mail.nagoya-u.ac.jp}
\and
College of Computer Science and Engineering, Shandong University of Science and Technology, China\\
%\email{housz@sdust.edu.cn}
\email{jiangshuai@hiroshima-u.ac.jp}
}
% \fi

%-------------------------------------------------------
\maketitle             
\begin{abstract}
The successful application of large pre-trained models such as BERT in natural language processing has attracted more attention from researchers. Since the BERT typically acts as an end-to-end black box, classification systems based on it usually have difficulty in interpretation and low robustness. This paper proposes a visual interpretation-based self-improving classification model with a combination of virtual adversarial training (VAT) and BERT models to address the above problems. Specifically, a fine-tuned BERT model is used as a classifier to classify the sentiment of the text. Then, the predicted sentiment classification labels are used as part of the input of another BERT for spam classification via a semi-supervised training manner using VAT. Additionally, visualization techniques, including visualizing the importance of words and normalizing the attention head matrix, are employed to analyze the relevance of each component to classification accuracy. Moreover, brand-new features will be found in the visual analysis, and classification performance will be improved.
Experimental results on Twitter's tweet dataset demonstrate the effectiveness of the proposed model on the classification task. Furthermore, the ablation study results illustrate the effect of different components of the proposed model on the classification results.

\end{abstract}

%-------------------------------------------------------
\section{Introduction}
\label{sec:intro}

Deep learning is a machine learning technique that has been widely applied in various fields, such as natural language processing (NLP) \cite{floridi2020gpt,li22transformer}, recommendation systems \cite{li2018capturing,marappan2022movie}, and prediction tasks \cite{li2022efficient,li2019multi}. In the field of spam email classification, deep learning models such as Recurrent Neural Networks, especially the Long Short-Term Memory \cite{zhang2019multi} and Gated Recurrent Unit \cite{li2021api}, have made significant progress in classifying emails and identifying spam. %Although each model has advantages and disadvantages, their performance can be improved by adjusting hyperparameters, preprocessing data, and model ensembling.

Finding suitable labels for domain-specific classification models trained using deep learning is challenging for researchers. Previous research has shown that sentiment analysis can be used in combination with pre-trained models for spam detection in tweets, as spammers often use emotional expressions to increase users' trust in their messages \cite{vosoughi2018spread}. However, determining effective tags for other types of social media content remains a challenge in this field \cite{farias2016irony}. In recent years, large pre-trained language models such as the BERT have achieved high accuracy when fine-tuning supervised tasks \cite{dai2015semi}. Additionally, some past work has partly studied the learning of linguistic features and examined the internal vector by probing the classifier \cite{radford2018improving}.

This paper uses spam detection as an example scenario. First, a fine-tuned BERT model is used for the sentiment classification of tweet texts. The sentiment label is then input for another BERT model to determine whether it is a spam tweet. In the training process of this model, a semi-supervised training approach using virtual adversarial training (VAT) is introduced to improve accuracy and system robustness. Ablation experiments demonstrate its effectiveness. Secondly, relevant tools are used to interpret the internal workings of the BERT model. By comparing various models used in the experiment and visualizing word importance attribution, the contribution of each token in each layer, and the attention matrices of each layer, the reasons for the accuracy differences between different models are found, explaining the models. Furthermore, more suitable URL tags are identified through internal analysis of the model. Further training of the model results in system improvement, as demonstrated by experiments. The main contributions of this paper are as follows:

\begin{itemize}
\item {\bf A BERT-based model for semi-supervised learning}: The first BERT is employed for text sentiment classification. The obtained sentiment tags are combined with the text in another BERT for spam classification via VAT.
\item {\bf Self-improved visual interpretation}: Word attention scores are analyzed with visual interpretation tools, and the parts with high feature weights are used to improve the system's classification performance.
\item {\bf Performance improvement}: Experimental results and ablation studies on the Twitter tweets dataset have demonstrated the effectiveness of the proposed model for spam classification in a semi-supervised learning task.

\end{itemize}

%-------------------------------------------------------
\section{Related Work}
\label{sec:related}

\subsection{Spam Detection}

Major social media sites (e.g., Twitter, Facebook, Quora) face a massive dilemma as many fall victim to spam. This information induces users to click on malicious links or uses bots to spread false news, seriously adding to the chaos in the Internet space. In recent years, many studies have been on spam detection for tweets, and many suitable and outdated features have been summarized \cite{kabakus2017survey}. Many studies have shown that sentiment analysis technology can enhance the differentiation of spam tweets \cite{antonakaki2021survey}. Therefore, many studies have used traditional machine learning methods to detect spam tweets based on sentiment features \cite{perveen2016sentiment,monica2020detection,saumya2018detection}. In \cite{rodrigues2022real}, the authors use several machine learning and deep learning techniques for sentiment analysis and spam detection to detect spam tweets in real time. However, the author does not combine the two techniques but performs spam detection and sentiment analysis on tweets separately. In \cite{ahsan2022spams}, the authors use an LDA model to find the sentiment and topic of tweets, suggest features that identify spam tweets more accurately than previous methods, and predict how widely spam spreads on Twitter. In \cite{heidari2020using}, the author first used the pre-trained BERT model to perform sentiment analysis on tweets, extracting various sentiment features. Then, an unsupervised GloVe model was used for Twitter bot detection, resulting in high accuracy. In addition, adversarial training has also been widely used in spam detection tasks. For example, in \cite{10.1145/1014052.1014066}, the author utilized several adversarial strategies to enhance the spam classifier and achieved good results, laying the foundation for adversarial training in classification tasks. \cite{9076796} used an attention mechanism for movie review spam detection and employed GAN models for adversarial training, achieving state-of-the-art results.

\subsection{Model Interpretability}

In today's era of widespread use of deep learning and neural network technology, the demand for their interpretability is also gradually increasing. Such models are usually black boxes in their organizational structure, where users input specific information into the model and can obtain specific outputs. However, the model still needs to answer how the outputs are obtained. Model interpretability aims to transform black-box models into white-box models so that users can understand why the model makes relevant predictions and identify ways to improve its validity. In addition, it eliminates ethical issues when AI models are used on a large scale in society.

\begin{figure}[t]
\centering
\includegraphics[width=1\linewidth]{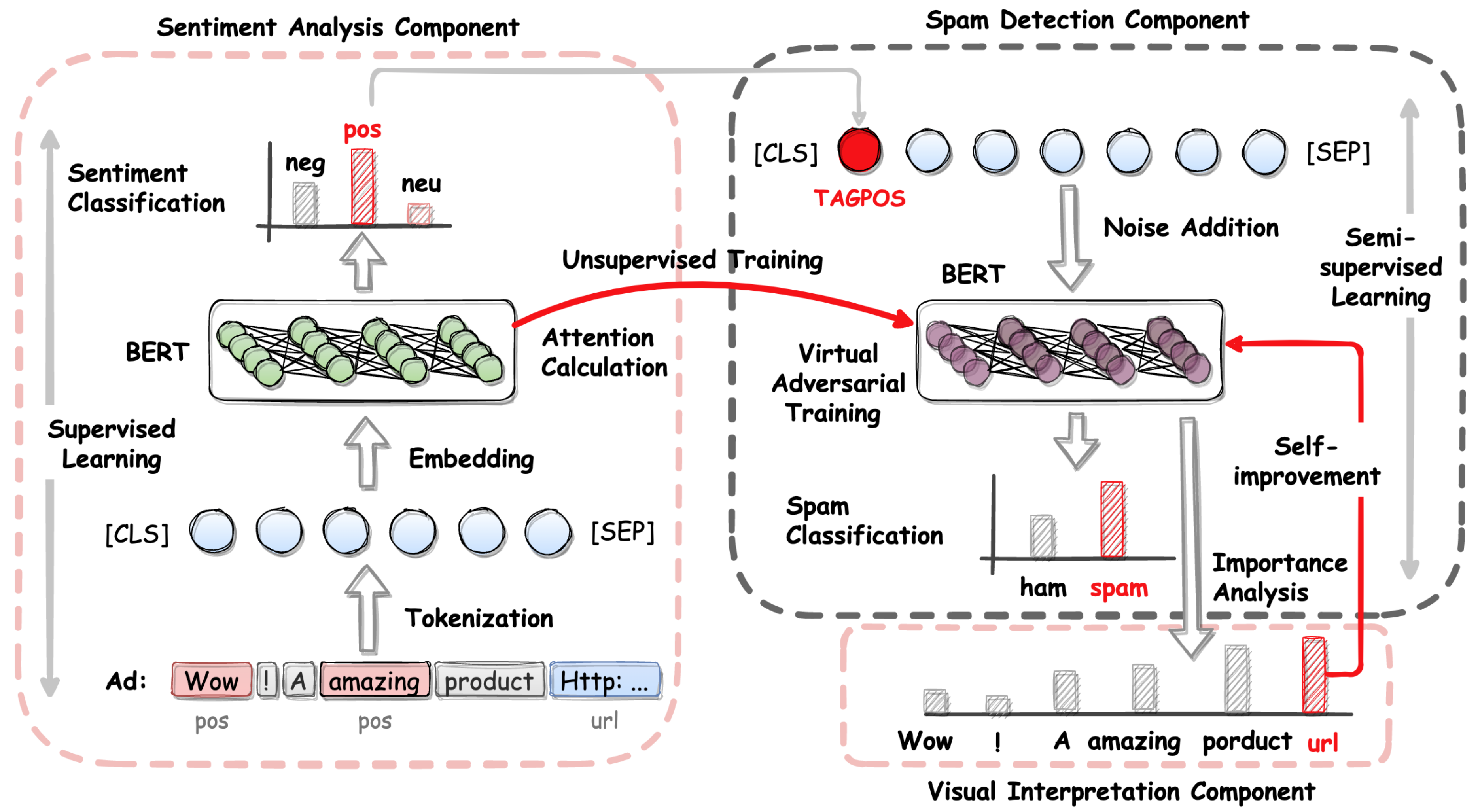}
\caption{Overview architecture of the proposed spam classification system. First, a fine-tuned BERT (the Sentiment Analysis Component) is used as a classifier to classify the sentiment of the text. Then, the predicted sentiment classification labels are used as part of the input of another BERT (the Spam Detection Component) for spam classification via a semi-supervised training manner using VAT.
Additionally, visualization techniques, including visualizing the importance of words and normalizing the attention head matrix, are employed to analyze the relevance of each component to classification accuracy. Moreover, brand-new features are detected in the Visual Analysis Component. Finally, the classification system can be self-improved via the newly imported.}
\label{fig1}
\end{figure}

Interpretability on machine learning models has long been proposed, such as SHAP \cite{lundberg2017unified} and LIME \cite{ribeiro2016should}. The SHAP (SHapley Additive exPlanations) model produces a prediction value for each prediction sample, and the SHAP value is the value assigned to each feature in that sample. Lime (Local Interpretable Model-Agnostic Explanations) is an approach that uses a trained local proxy model to explain individual samples. However, when dealing with large pre-trained deep learning models with hundreds of hidden states, the situation becomes different, and simple local explanations of the model are difficult to fit. The BERT model, for example, introduced the attention mechanism \cite{bahdanau2014neural}, which became a very successful neural network component but increased the difficulty of interpreting the model. Clark et al. \cite{clark2019does} strongly emphasize analyzing the attention head in BERT. They studied its behavior and directly extracted sentence representations from the BERT model without fine-tuning. They discovered that the attention head exhibits recognizable patterns, such as focusing on a fixed position offset or paying attention to the entire sentence.

%-------------------------------------------------------
\section{Model}

To identify spam tweets, we used the public dataset “Spam Detection on Twitter” \cite{Yash} posted by YASH, which contains 82,469 legitimate tweets and 97,276 spam tweets. To classify the sentiment of tweets, we utilized the Sentiment140 dataset \cite{go2009twitter}, which comprises 1.6 million tweets, half positive and half negative. The model architecture is shown in Fig. \ref{fig1}. First, the tweets will be tagged with sentiment labels after a BERT model fine-tuned by the Sentiment140 dataset, the Sentiment Analysis Component. Then, after two fully connected layers, the Spam Detection Component will output whether the tweet is spam. Semi-supervised learning and VAT are used here to improve the training accuracy. We utilize the Twitter dataset as unlabeled data for semi-supervised learning in sentiment analysis. Finally, the interpretation method will interpret the models.

% \subsection{Model Structure}

\subsection{Sentiment Analysis Component}

Sentiment analysis of tweets helps to comprehend public opinion on topics prevalent on social media. Twitter's usage has increased as users share news and personal experiences. Hence, analyzing tweet sentiment is crucial. Despite its popularity, sentiment analysis of tweets is challenging due to the 280-character limit and irregularities in tweets (e.g., spelling variations and abbreviations).

BERT is a pre-trained deep bidirectional transformer, a powerful model for language understanding. We employed BERT for sentiment polarity classification using the Sentiment140 dataset. This dataset contains tweet content and is for a binary sentiment classification task. It includes 1.6 million tweets collected from the Twitter API and annotated as negative (0) or positive (4), making it useful for sentiment detection. Unlike previous work \cite{heidari2020using} by Heidari et al., who used the SST-2 movie review dataset for training, the Sentiment140 dataset is in the same domain as the target task, thus leading to improved accuracy in sentiment analysis of tweets.

% \begin{figure}[htbp]
% \centerline{\includegraphics[width=0.8\linewidth]{figs/fig2.png}}
% \caption{Overall procedures for BERT \cite{devlin2018bert}}
% \label{fig2}
% \end{figure}

The Sentiment Analysis Component categorizes tweets as positive, neutral, or negative. BERT is trained on 1.6 million tweets. After the tokens are input into the model, the model performs word embedding processing. Among the 12 hidden layers in the BERT model, the next layer's multi-head attention calculates the attention scores of each word in the previous layer. The BERT model's training results will be presented in the next section.

This component finally extracts sentiment features from the text of tweets through fine-tuned BERT. At the end of the model, the softmax layer outputs the sentence's sentiment polarity value score $X$ ($0< X< 1$). Then it divides the sentence into three categories: positive, neutral, and negative, according to this value. If $X<0.3$, the tweet sentiment is negative. If $X> 0.7$, the tweet sentiment is positive. The rest are neutral.

\subsection{Spam Detection Component}

The Spam Detection Component also uses the fine-tuned BERT to determine if the tweet is spam. We use the \emph{SpamDetectionOnTwitter} dataset in \cite{Yash} to learn whether a tweet is a spam. To embed emotional features into tweets and better explain the model, we add sentiment tags into each piece of data, namely \emph{TAGPOS}, \emph{TAGNEU}, and \emph{TAGNEG}, and add these three tags to the dictionary of the BERT model. After exporting the BERT model, it includes two fully connected layers. The final layer with softmax will output the final result indicating whether the tweet is spam.

% \begin{figure}[htbp]
% \centerline{\includegraphics[width=0.9\linewidth]{figs/fig3.png}}
% \caption{Example of dataset structure.}
% \label{fig3}
% \end{figure}

At the same time, we also use several adversarial learning methods for training enhancement to find the best one. Adversarial training can be summarized as the following max-min formula:

\begin{equation} \label{at_eq1}
\min\limits_{\theta}\mathbb{E}_{(x,y)\sim\mathcal{D}}{[\max\limits_{||\delta||\le\varepsilon}L(f_\theta(X+\delta),y)]}.
\end{equation}

The inner layer (in square brackets) is a maximization, where $X$ represents the input representation of the sample, $\delta$ represents the perturbation superimposed on the input, $f_\theta()$ is the neural network function, $y$ is the label of the sample, and $L(f_\theta(X+\delta),y)$ represents the loss obtained by superimposing a disturbance $\delta$ superimposed on the sample $X$. $\max(L)$ is the optimization goal, that is, to find the disturbance that maximizes the loss function. The outside minimization refers to finding the most robust parameters $\theta$, such that the predicted distribution conforms to the distribution of the original dataset.

The Fast Gradient Method (FGM) is implemented by $L2$ normalization, which divides the value of each gradient dimension by the $L2$ norm of the gradient. Theoretically, $L2$ normalization preserves the direction of the gradient.
\begin{equation} \label{fgm_eq1}
\delta=\epsilon\bullet\left({g/|\left|g\right||}_2\right).
\end{equation}
Among them, $g=\mathrm{\nabla\ _X}\left(L\left(f_\theta\left(X\right),y\right)\right)$ is the gradient of the loss function $L$ with respect to the input $X$. Unlike a normal FGM that only performs iteration once, PGD performs multiple iterations to find the optimal perturbation. Each iteration projects the disturbance into a specified range each time a small step is taken. The formula of the loss function in step $t$ in PGD is shown as follows:
\begin{equation} \label{pgd_eq1}
g_t=\mathrm{\nabla}X_t\left(L\left(f_\theta\left(X_t\right),y\right)\right).
\end{equation}
Although PGD is simple and effective, there is a problem that it is not computationally efficient. Without adversarial training, $m$ iterations will only have $m$ gradient calculations, but for PGD, each gradient descent must correspond to the $K$ steps of gradient boosting. Therefore, PGD needs to do $m(K+1)$ gradient calculations compared with the method without adversarial training.
In VAT, the loss function for adversarial training can be expressed as \cite{miyato2018virtual}:
\begin{equation} \label{eq1}
L_{adv}(x_l, \theta) := D [q(y|x_l), p(y|x_l + r_{adv}, \theta)],
\end{equation}
\begin{equation} 
\label{eq2}
r_{adv} := \mathop{\arg\max}\limits_{r;||r||\leq\epsilon} D [q(y|x_l), p(y|x_l + r, \theta)]\text{,}
\end{equation}
where $D[p, q]$ is a non-negative function that measures the divergence between two distributions $p$ and $q$. The function $q(y|x_l)$ is the true distribution of the output label, which is unknown. This loss function aims to approximate the true distribution $q(y|x_l)$ by a parametric model $p(y|x_l, \theta)$ that is robust against adversarial attack to labeled input $x_l$. A "virtual" label generated by the $p(y|x, \theta)$ probability is used in VAT to represent the user-unknown $p(y|x, \hat\theta)$ label, and the adversarial direction is calculated based on the virtual label. Unlabeled input $x_{ul}$ and labeled input $x_l$ will be unified as $x_\ast$. The formula is calculated as follows:

\begin{equation} \label{eq3}
\text{LDS}(x_\ast, \theta) := D [p(y|x_\ast,\hat\theta), p(y|x_\ast + r_{qadv}, \theta)],
\end{equation}
\begin{equation} \label{eq4}
r_{qadv} := \mathop{\arg\max}\limits_{r;||r||_2\leq\epsilon} D [p(y|x_\ast,\hat\theta), p(y|x_\ast + r)],
\end{equation}
 where the loss function of LDS$(x_\ast, \theta)$ indicates the virtual adversarial perturbation. This function can be considered a negative indicator of the local smoothness of the current model at each input data point $x$. A reduction in this function would result in a smoother model at each data point.

\subsection{Visual Interpretation Component}

In this part, we begin by creating visual representations of the significance of individual words in differentiating between spam and non-spam content. Additionally, we normalize the attention head matrix to visualize all attention matrices and identify distinctions between various models, thereby demonstrating the efficacy of our proposed model. We will delve into further details in the following part, accompanied by relevant examples.

\subsubsection{Word Importance Attribution}
Integrated Gradients \cite{sundararajan2017axiomatic} are used to compute attributions concerning the \emph{BertEmbeddings} layer to obtain the importance of words. In simple terms, Integrated Gradients define the attribution of the $i^{th}$ feature of the input as the path integral of the straight line path from the baseline $x'_i$ to the input $x_i$ from \cite{sundararajan2017axiomatic}:

\begin{equation} \label{eq11}
\text{IG}_i(x) ::= (x_i-x'_i)\cdot\int^1_{\alpha=0}\frac{\partial F(x'+\alpha(x-x'))}{\partial x_i}\text d\alpha,
\end{equation}
where $\frac{\partial F(x)}{\partial x_i}$ is the gradient of $F$ along the $i^{th}$ dimension at input $x$ and baseline $x'$. In the NLP task described in this paper, we use the zero vector as the baseline.

\subsubsection{Attribution in Attention Matrix}
We visualize the attention probabilities of 12 attention heads in all 12 layers, totaling 144. It represents the softmax normalized dot product of key and query vectors. In \cite{clark2019does}, it is an essential indicator, indicating how related a token is to another token in the text.

%-------------------------------------------------------
\section{Experiments}
\label{sec:exp}

To demonstrate the effectiveness of the proposed model, this section first provides empirical evidence through ablation experiments, demonstrating the effectiveness of the relevant components, including the sentiment analysis component and the adversarial training component. Secondly, visualization tools are used to analyze the model's interpretability to identify the reasons for the effectiveness of the relevant components. Finally, using the above analysis, more suitable labels are identified, and the system is further trained to achieve improvements.

\subsection{Dataset}
\subsubsection{Spam Dataset}

We use the SpamDetectionOnTwitter dataset in \cite{Yash} to learn whether a tweet is spam. This dataset contains 82,469 legitimate tweets and 97,276 spam tweets. Each tweet is tagged with user\_id, tweet\_id, tweet\_text, time, and spam\_label to show whether it is a spam tweet. Here we only select the text and spam\_label for training. We select 68,919 legitimate tweets and 58,866 spam tweets as the training set, and the rest is divided into a validation set and a test set.

\subsubsection{Sentiment Dataset}
We used the Sentiment140 dataset \cite{go2009twitter} as the training dataset for the part of the tweet sentiment polarity analysis component. This dataset contains 1.6 million sentiment-labeled tweets, half positive and half negative, and each tweet is accompanied by tweet\_id, time, username, and tweet\_text. Similar to the last part, only the tweet\_text and spam\_label are selected for training at this stage. We select 1.46 million tweets as the training set, and the rest are divided into the validation and test sets.

\subsection{Hyperparameter Setting}

In both Sentiment Analysis Component and Spam Detection Component, we used the bert-base-multilingual-cased model, which is Google's new and recommended BERT model. We set the batch size to 16 and the dropout rate to 0.1 in both Sentiment Analysis and Spam Detection Components. The learning rates of the Adam optimizer are 2e-5 in the sentiment part and 1e-5 in the spam part. According to the size of the two datasets, we set the steps of the sentiment part as 10000, while 1000 in the spam part. All experiments used Pytorch version 1.13.1, bert4torch 0.2.4, and Captum 0.6.0.

\subsection{Spam Detection}

% Figure \ref{fig4a} and figure \ref{fig4b} demonstrate the effectiveness of the Sentiment Analysis Component proposed by the model. The x-axis is the training steps. The y-axis is the test set accuracy and training loss, respectively. Figure \ref{fig4a} shows the test set accuracy, which achieved an accuracy of 78.83 \% at the end of the training, and the training set loss in Figure \ref{fig4b} also proves that the model has converged.

% \begin{figure}[htbp]
%     \centering
%     \subfigure[]{
%     \includegraphics[width=0.4\textwidth]{figs/fig4a.pdf}
%     \label{fig4a}
%     }
%     \subfigure[]{
%     \includegraphics[width=0.4\textwidth]{figs/fig4b.pdf}
%     \label{fig4b}
%     }
% \caption{Training process of Sentiment Analysis Component.}
% \label{fig4}
% \end{figure}

\begin{figure}[b]
\centering
\includegraphics[width=0.5\textwidth]{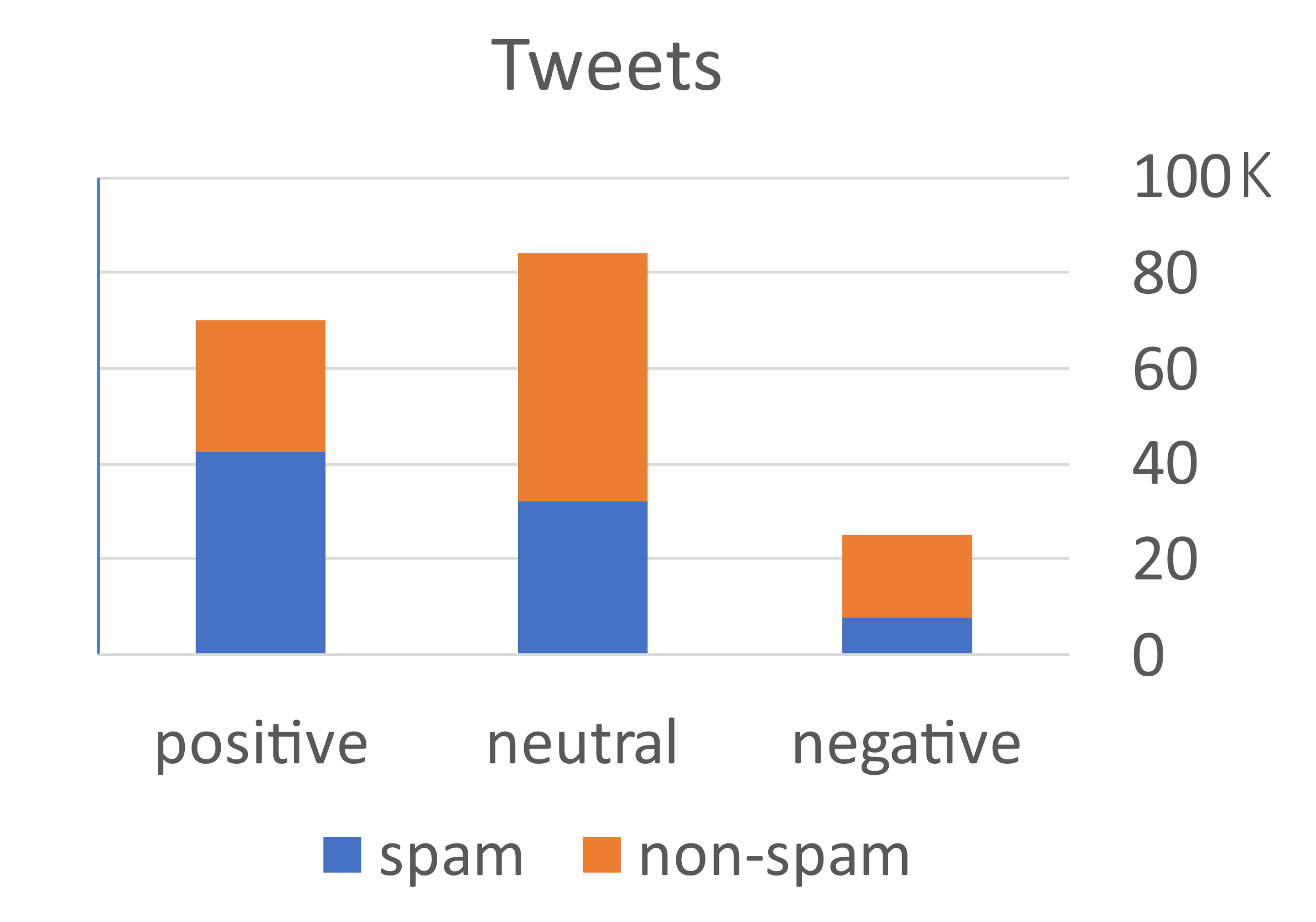}
\vspace{-15pt}
\caption{Sentiment distribution in spam dataset.}
\label{fig5}
\end{figure}

After fine-tuning, Sentiment Analysis is performed on the existing spam dataset, and the sentiment distribution of the dataset can be seen, as shown in Fig. \ref{fig5}. The number of spam tweets with positive sentiment is the largest, followed by neutral sentiment and the least negative sentiment, with 42547, 32291, and 7629, respectively. In the non-spam tweets, the tweets with neutral sentiment are the most, and the positive and negative sentiment is both less, among which the negative sentiment is the least, the numbers 27619, 52049, and 17606, respectively.

% \begin{figure}[htbp]
%     \centering
%     \subfigure[Distribution of spam labels by sentiment category.]{
%     \includegraphics[width=0.48\textwidth]{figs/fig5a.pdf}
%     \label{fig5a}
%     }\hspace{-2mm}
%     \subfigure[Distribution of sentiment labels by spam category.]{
%     \includegraphics[width=0.48\textwidth]{figs/fig5b.pdf}
%     \label{fig5b}
%     }
% \caption{Sentiment distribution in spam dataset}
% \label{fig5}
% \end{figure}

% Figure \ref{fig6a} and figure \ref{fig6b} show the proposed models' training data. The x-axis shows the number of epochs, while the y-axis shows the accuracy of the test set and the training loss. In the later stages of training, accuracy rate and loss gradually become stable, indicating that the model has converged.
To show the effectiveness of the proposed model, we performed ablation experiments on the model proposed in this paper using the same training parameters and random seed. Since PGD is an evolutionary algorithm of FGM, here we omit the experiment of FGM and only keep PGD. The results are shown in Table \ref{table1}. It can be seen that the precision of the proposed model is the highest, proving its effectiveness. This part only analyzes the model's effectiveness from the model experimentation perspective. Although the proposed model has achieved the highest accuracy and recall, as a black box model, we cannot know the reasons for the differences in the experiment results inside the different models. This issue leads us to introduce model interpretability (or XAI). The next part will study this issue in depth.

% \begin{figure}[htbp]
%     \centering
%     \subfigure[]{
%     \includegraphics[width=0.4\textwidth]{figs/fig6a1.pdf}
%     \label{fig6a}
%     }
%     \subfigure[]{
%     \includegraphics[width=0.4\textwidth]{figs/fig6a1.pdf}
%     \label{fig6b}
%     }
% \caption{Test accuracy and training loss of the proposed model.}
% \label{fig6}
% \end{figure}

\vspace{-15pt}
\begin{table}[htbp] 
\setlength\tabcolsep{10pt}
\caption{Experiment results of all models.}
\label{table1}
\centering
% \scalebox{0.91} 
{

% \resizebox{\linewidth}{!}{

\begin{tabular}{l l l l l}
\hline
& Precision \  & Accuracy \ & Recall \  & F1 Score     \\ \hline
BERT                & 76.06    & 75.32     & 79.95     & 77.96  \\
BERT+Sentiment      & 79.14    & 77.43	   & 79.65     & 79.39  \\ 
BERT+Sentiment+PGD \   & 83.21    & 76.83	   & 72.10     & 77.25  \\
BERT+Sentiment+VAT \  & 82.61    & 76.68     & 79.54     & 78.47  \\ \hline
\makecell[l]{Proposed Model\\(1 dense layer)}
& 82.58    & 77.81     & 75.41     & 78.83 \\
Proposed Model
& 85.97    & 77.60	& 74.92     & 78.49  \\ \hline
\end{tabular}
% }

}
\end{table}
\vspace{-25pt}
\subsection{Visual Interpretation}

In this section, we use the Captum \cite{kokhlikyan2020captum} tool to perform visual interpretability analysis on the BERT model in the Spam Detection Component of all six models in the previous chapter. This tool is a Pytorch-based model interpretation library released by Facebook. The library provides interpretability for many new algorithms (such as ResNet, BERT, and some semantic segmentation networks), helping everyone better understand the specific features, neurons, and neural network layers that affect the model's prediction results. For text translation and other problems, it can visually mark the importance of different words and use a heat map to display the correlation between words.

Here, we first visualize the importance of words to distinguish which words play a role in judging spam or not. Then, we visualize all the attention matrices to find the differences between different models to prove the effectiveness of the proposed model. Here is an example of the tweet “\emph{19 year old genius shares Twitter tool free. Nice guys rock! http://ow.ly/Ul1t}”, a spam tweet with positive sentiment.

\subsubsection{Word Importance Attribution}

\begin{figure}[htbp]    
    \subfigure[BERT model]{
    \includegraphics[width=0.98\textwidth]{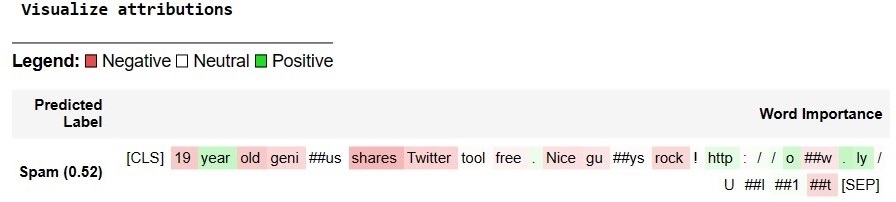}
    \label{fig7a}
    }
    
    \subfigure[Proposed model]{
    \includegraphics[width=0.98\textwidth]{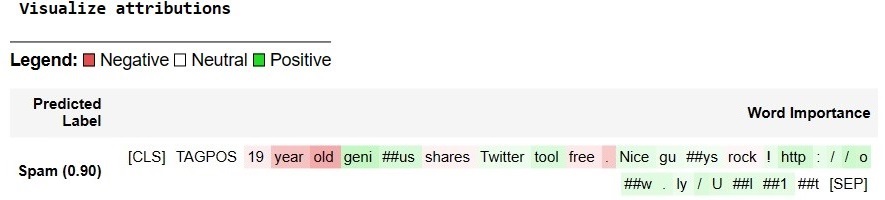}
    \label{fig7b}
    }
    \vspace{-15pt}
    \caption{Word importance attribution.}
    \label{fig7}
\end{figure}

With the formula (\ref{eq11}), we can obtain the Word Importance Attribution of the input sentence, as shown in Fig. \ref{fig7}. 
In this case, the actual label of the input sentence is "Spam."
This figure's ``Predicted Label'' represented the model output result and predicted probability. 
The rightmost is a visual explanation of the contribution value of the input sentence, green represents a positive contribution to the Predicted Label, and red represents the opposite. Additionally, the deeper the color, the higher the level of contribution, and vice versa.

It can be seen from the figure that when the sentiment tag is not added, only the word \emph{year} is a positive contribution, so the final probability is only 0.52, and the model is difficult to distinguish. After adding the sentiment tag, although the contribution of the tag is less, the contribution of many words, especially the URL part, is significantly improved, thereby increasing the final probability.

\vspace{-15pt}
\vspace{-20pt}
\subsubsection{Attribution in Attention Matrix}

The order of the following figures is simple BERT model and the proposed model. The $x$-axis and $y$-axis of the matrix are tokens, and each cell represents the attention score between different tokens, that is, the degree of attention obtained from the weight of attention. The brighter the cell, the higher the attention score, or level of attention, from the token on the $x$-axis to the token on the $y$-axis, and vice versa.

\begin{figure}[htbp] 
    \centering
    \includegraphics[width=1\linewidth]{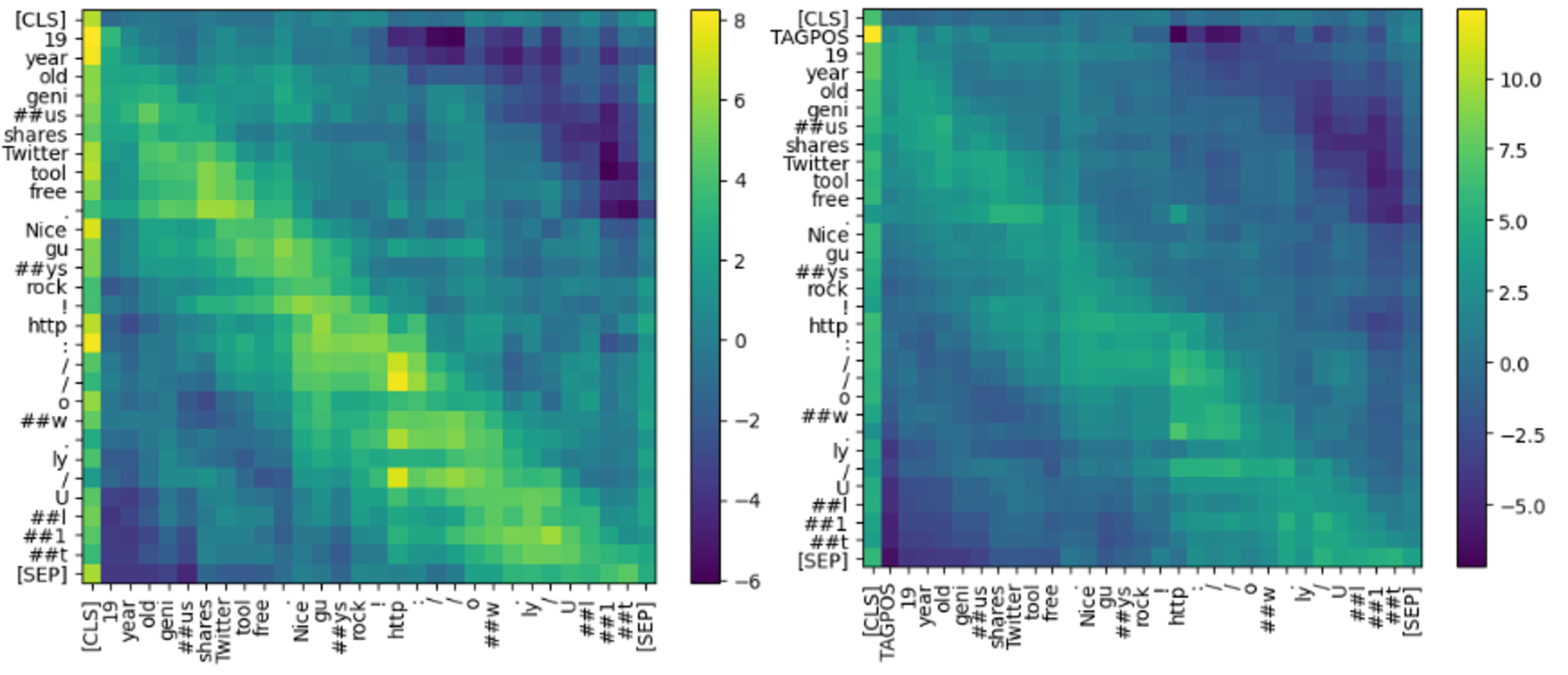}
    \caption{Attention Head 3-6 of BERT and the proposed model.}
    \label{fig10}
%\end{figure}
%\begin{figure}[htbp] 
    \centering
    \includegraphics[width=1\linewidth]{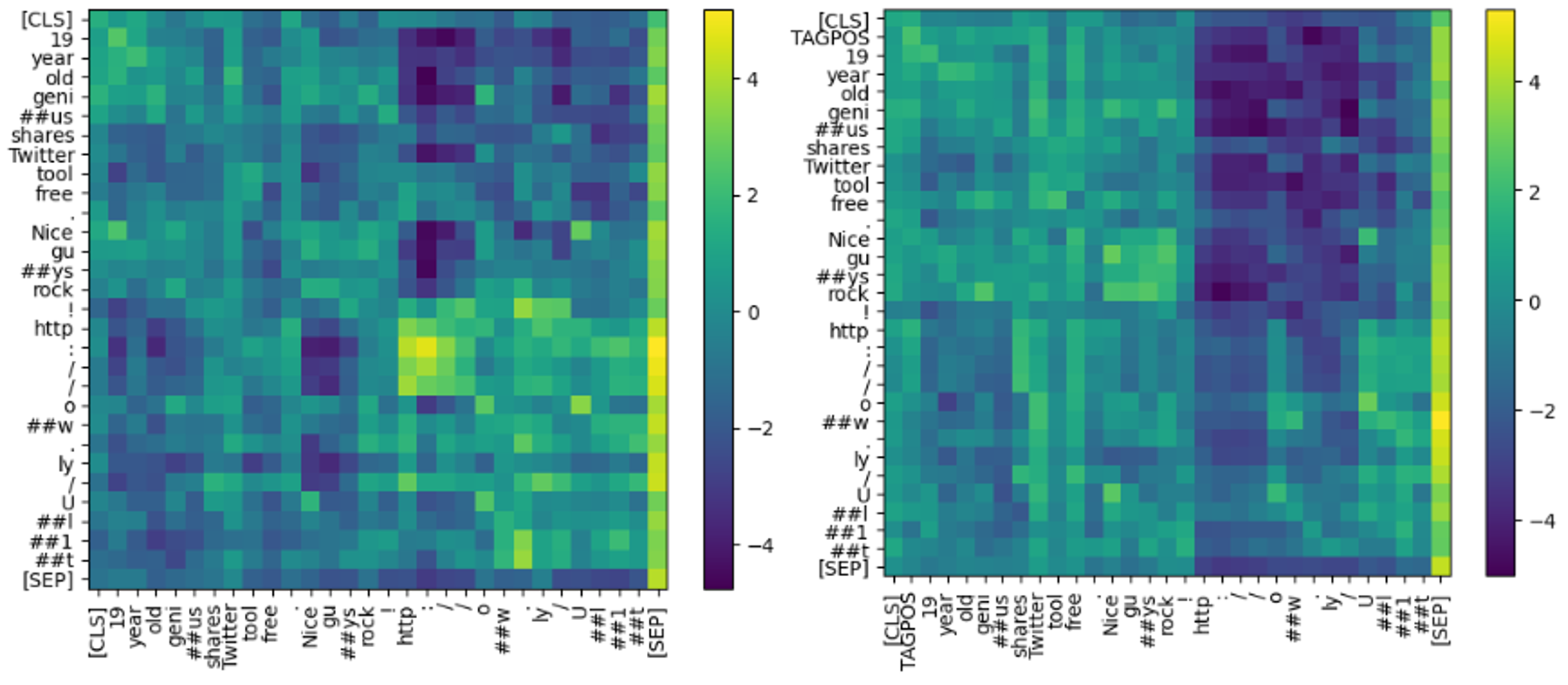}
    \caption{Attention Head 9-12 of BERT and the proposed model.}
    \label{fig14}
%\end{figure}
%\begin{figure}[ht] 
    \centering
    \includegraphics[width=1\linewidth]{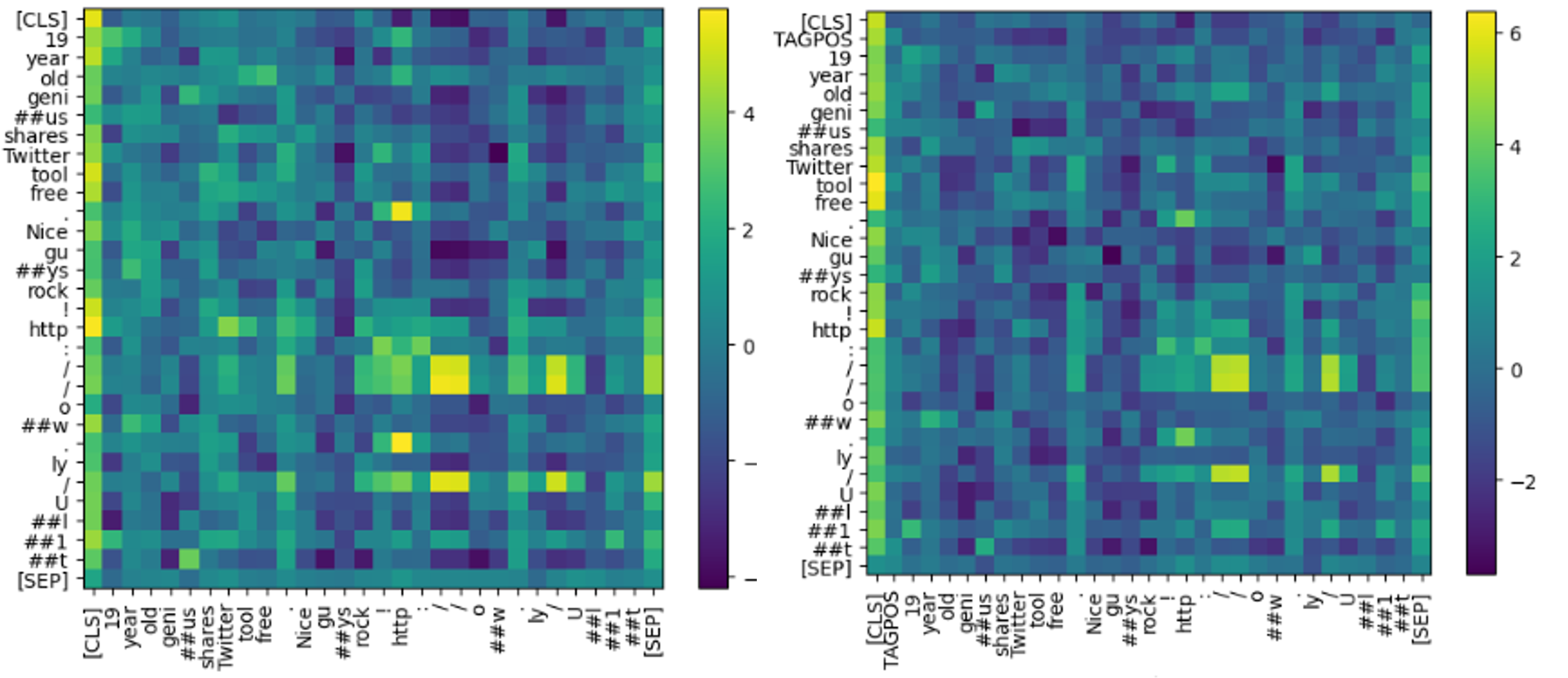}
    \caption{Attention Head 2-12 of BERT and the proposed model.}
    \label{fig16}
\end{figure}

In most attention heads, the overall trend of words is to pay more attention to themselves or the next word. Still, in Head 2-5, 3-3, 3-4, 3-6, 3-8, 3-12, and 4-2, the [CLS] token will pay more attention to the added sentiment tag, which we can see the brightest point appears in the upper left corner, such as Head 3-6 (Fig. \ref{fig10}). According to \cite{devlin2018bert}, each sequence's initial [CLS] token is used as the sentence representation in a labeled classification task. Therefore, in the Spam Detection Component of this paper, the [CLS] token represents whether the sentence is spam. Therefore, these results show the validity of the sentiment tag. Entering layer 9, it can be seen that, as shown in Fig. \ref{fig14}, the attention matrix divides the token into two parts, the text, and the URL. This phenomenon is more evident in Head 9-12, and the attention matrix is divided into four prominent parts, especially in the model after adding the adversarial training method.

Other examples can also demonstrate the effectiveness of the proposed model. In Head 2-12, shown in Fig. \ref{fig16}, we can see the part of the URL that pays more attention to \emph{http://}, indicating that the model has detected the URL.

\begin{figure}[ht] 
    \centering
    \centerline{\includegraphics[width=1\linewidth]{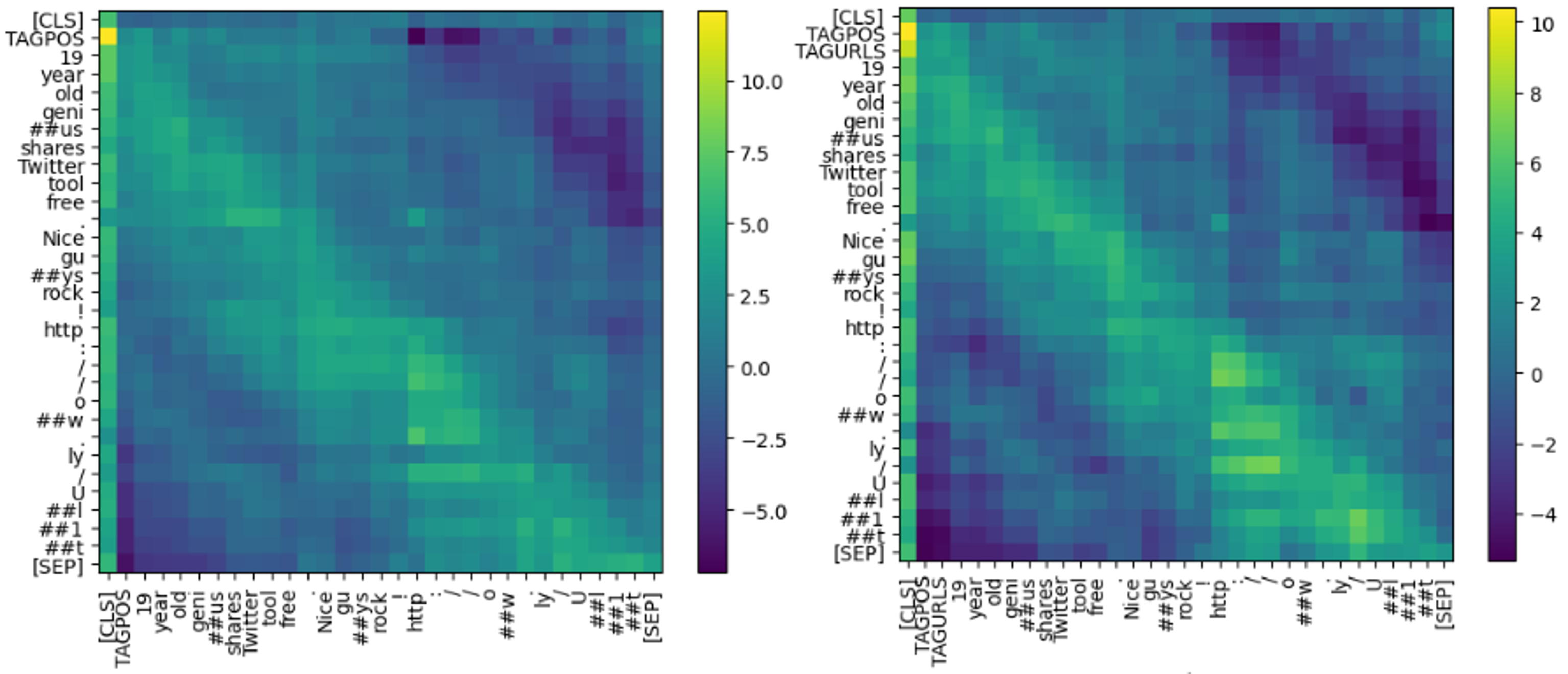}}
    \caption{Attention Head 3-6 of the proposed model and improved model.}
    \label{fig17}
%\end{figure}
%\begin{figure}[htbp] 
    \centering
    \centerline{\includegraphics[width=1\linewidth]{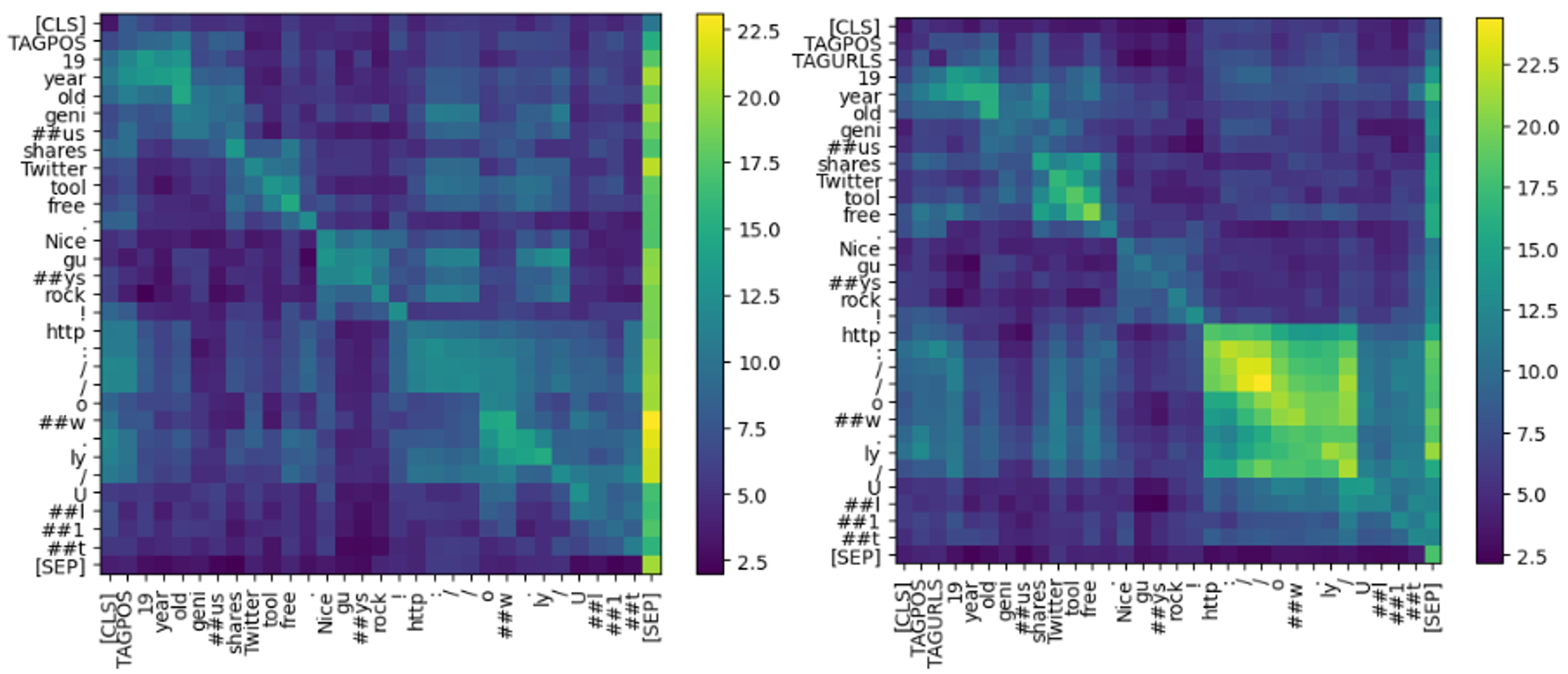}}
    \caption{Attention Head 9-12 of the proposed model and improved model.}
    \label{fig18}
\end{figure}

% \begin{figure}[htbp] 
%     \centering
%     \centerline{\includegraphics[width=\linewidth]{figs/fig19.png}}
%     \caption{Layer 9 of the proposed model and modified model}
%     \label{fig19}
% \end{figure}

\subsection{Model Improvement}

Based on the analysis in the previous section, we found that the URL part can be detected in some attention heads. Inspired by this, we operate similarly to the sentiment tags above for the URL part, using a regular expression to extract the URL part as separate data labels. To distinguish between short and long links, we set the URL label with less than 24 characters as \emph{TAGURLS}, indicating short URL links, and the rest as \emph{TAGURLL}. We add these labels after the sentiment label. We used the optimized data to retrain the proposed model, resulting in the modified model, as shown in Table \ref{table2}. We found that after adding the URL tag, the accuracy, F1 score, and precision of the modified model were improved.

\begin{table}[ht] 
\setlength\tabcolsep{10pt}
\caption{Experiment results of the proposed model and modified model}
\label{table2}
\centering
{
\begin{tabular}{lllll}
\hline
    & Precision \   & Accuracy \ & Recall \   & F1 Score \       \\ \hline
Proposed Model \ & 85.97     & 77.60	& 74.92     & 78.49  \\ \hline
Improved Model & \textbf{87.13}   & \textbf{77.85} & 75.78  & 78.88  \\ \hline
\end{tabular}
}
\end{table}

Next, we analyze the BERT model using interpretability methods and directly examine the corresponding attention head of the original model. In Fig. \ref{fig17}--\ref{fig18}, the left part is the proposed model, and the right is the improved model. In Fig. \ref{fig17}, Head 3-6 is the same as before, with [CLS] paying the most attention to the sentiment tag, but the URL tag has become the second most attention token, which to some extent, proves the validity of the URL tag. In Fig. \ref{fig18}, Head 9-12, the URL part is more prominent than the original model, indicating that the model is paying more attention to the URL part, proving the URL tag's reliability for improving precision and accuracy.

%-------------------------------------------------------
\section{Conclusion}
\label{sec:conclusion}

The main contributions of this paper can be summarized in two points. First, using sentiment analysis and adversarial training methods, we proposed a new model for spam detection, which is better than the traditional models. Secondly, applying the visual interpretability analysis method to the model, we studied the principle of internal classification of the model, found the reasons for the difference in precision in different models, and proved the effectiveness of the proposed model at the same time, further improving its performance.

The large-scale pre-trained BERT model based on VAT can be extended to other tasks. The attention mechanism can analyze the in-depth features with heavy weights, and these features can effectively improve the accuracy and precision of the task. In future work, we will utilize the VAT and visual interpretation method in other pre-trained language models (e.g., ALBERT, XLNET) to further improve the performance of spam classification.

% \bibliographystyle{splncs04}
% \bibliography{refs}

\end{document}